\documentclass{article}
\usepackage{amsmath}

\usepackage{arxiv}

\usepackage[utf8]{inputenc} 
\usepackage[T1]{fontenc}    
\usepackage{hyperref}       
\usepackage{url}            
\usepackage{booktabs}       
\usepackage{amsfonts}       
\usepackage{nicefrac}       
\usepackage{microtype}      
\usepackage{lipsum}
\usepackage{graphicx}
\graphicspath{ {./images/} }

\title{SME-YOLO: A Real-Time Detector for Tiny Defect Detection on PCB Surfaces}

\author{
  Meng Han\\
  International School of Technology\\
  Henan University\\
  Zhengzhou, Henan ,China \\
  \texttt{2124030172@henu.edu.cn} \\
}

\begin{document}
\maketitle
\begin{abstract}
Surface defects on Printed Circuit Boards (PCBs) directly compromise product reliability and safety. However, achieving high-precision detection is challenging because PCB defects are typically characterized by tiny sizes, high texture similarity, and uneven scale distributions. To address these challenges, this paper proposes a novel framework based on YOLOv11n, named SME-YOLO (Small-target Multi-scale Enhanced YOLO). First, we employ the Normalized Wasserstein Distance Loss (NWDLoss). This metric effectively mitigates the sensitivity of Intersection over Union (IoU) to positional deviations in tiny objects. Second, the original upsampling module is replaced by the Efficient Upsampling Convolution Block (EUCB). By utilizing multi-scale convolutions, the EUCB gradually recovers spatial resolution and enhances the preservation of edge and texture details for tiny defects. Finally, this paper proposes the Multi-Scale Focused Attention (MSFA) module. Tailored to the specific spatial distribution of PCB defects, this module adaptively strengthens perception within key scale intervals, achieving efficient fusion of local fine-grained features and global context information. Experimental results on the PKU-PCB dataset demonstrate that SME-YOLO achieves state-of-the-art performance. Specifically, compared to the baseline YOLOv11n, SME-YOLO improves mAP by 2.2\% and Precision by 4\%, validating the effectiveness of the proposed method.
\end{abstract}

\section{Introduction}
As the core carrier of electronic devices, the Printed Circuit Board (PCB) plays a critical role in product reliability and safety\cite{1,2,3}. Surface defects on PCBs directly threaten the functionality of the entire system\cite{4}. In high-sensitivity fields such as transportation, aerospace, and healthcare, electrical anomalies caused by PCB defects can lead to system failures\cite{5}, potentially resulting in service interruptions, financial losses\cite{4}, or even casualties. Economically, missed detections before factory shipment often trigger large-scale product recalls, causing substantial consequential damages. With the continuous rise in global electrification and informatization, the demand for PCBs shows a long-term growth trend. Therefore, developing efficient and accurate methods for PCB surface defect detection is of significant theoretical and practical importance.

Early PCB defect detection relied on manual visual inspection\cite{6}. This approach is not only expensive and inefficient but also highly susceptible to external factors such as subjectivity, environmental conditions\cite{7}, and employee fatigue. Consequently, manual inspection struggles to meet the precision and efficiency demands of the electronics manufacturing industry\cite{8}. With the rapid development of computer vision, object detection technologies based on deep learning have gradually replaced manual inspection due to their low cost, high efficiency, and automation capabilities. Among them, the YOLO (You Only Look Once) series has become a mainstream choice in industrial inspection for its excellent balance between detection speed and accuracy\cite{9,10,11}.

Although YOLO models perform excellently in general object detection tasks, directly applying them to PCB surface defect detection faces several challenges: (1) Difficulty in detecting tiny objects: Most defects on PCB surfaces are spatially minute compared to the entire board, typically occupying only 0.1\% of the total area in high-resolution images. These features are easily lost during downsampling and feature fusion. (2) High similarity in local textures: The local structures of PCBs are highly repetitive, making it difficult for models to distinguish between normal structures and defects. (3) Uneven multi-scale distribution: Different types of defects (e.g., pinholes, scratches, block contamination) exhibit significant unevenness in spatial scales, requiring detectors to possess stronger multi-scale adaptability.

Existing improvements to YOLO for PCB defect detection have optimized feature enhancement, attention mechanisms, and multi-scale fusion. However, these methods share common limitations: (1) Sensitivity of IoU-based loss functions: These functions are highly sensitive to positional deviations in tiny objects, leading to insufficient localization precision for minute defects. (2) Limitations of common upsampling: Standard upsampling methods have limited capacity to express fine-grained textures when recovering spatial resolution, often smoothing or diluting boundary information of tiny objects. (3) Lack of optimization in attention mechanisms: Current attention mechanisms are not optimized for the specific scale distribution of PCB defects. While they improve multi-scale adaptability, they waste computational power on non-critical scales and fail to extract sufficient features at the critical tiny scales.

To address these issues, this paper proposes a lightweight and high-precision detection model named SME-YOLO (Small-target Multi-scale Enhanced YOLO), based on the YOLOv11n architecture. The main contributions are as follows:

\paragraph{1)Introduction of Normalized Wasserstein Distance Loss (NWDLoss).}The bounding box loss function is replaced with NWDLoss\cite{12}. Bounding boxes are modeled as 2D Gaussian distributions, and the Normalized Wasserstein Distance is used to measure the similarity between predicted and ground-truth boxes. This approach effectively mitigates the sensitivity of IoU to positional deviations in tiny objects, improving localization precision.

\paragraph{2)Optimization of upsampling with EUCB.}The Efficient Upsampling Convolution Block (EUCB)\cite{13} replaces the original upsampling module. By employing multi-scale convolutions, EUCB gradually recovers the spatial resolution of feature maps, enhancing the retention of edge and texture details for tiny objects.

\paragraph{3)Enhancement of multi-scale feature extraction with MSFA.}The Multi-Scale Focused Attention (MSFA) module replaces the C3k2 structure in the backbone network. This module reconstructs the convolution kernel configuration based on the scale distribution characteristics of PCB defects. It utilizes dual same-scale branches for implicit integration and removes redundant large kernels to reduce computational overhead, achieving an effective balance between multi-scale perception and computational efficiency.

Experiments on the PKU-Market-PCB dataset\cite{14}, curated by the Intelligent Robot Open Laboratory of Peking University, demonstrate that the proposed method achieves 97.0\% mAP with a low parameter count. This represents a 2.2\% improvement over the baseline YOLOv11n, validating the effectiveness of the proposed improvements.

\section{Relate work}
\subsection{Overview of PCB Surface Defect Detection Methods}
\label{sec:headings}
The technology for PCB surface defect detection has evolved from traditional image processing to machine learning, and finally to deep learning. Early detection methods primarily relied on traditional image processing techniques, including template matching, threshold segmentation, and edge detection. Hassanin et al.\cite{15}proposed an automated PCB detection method based on SURF features. By constructing a descriptor dictionary of component features and combining it with image registration, filtering, segmentation, and morphological operations, this method achieved accurate localization and identification of missing components. Chang et al.\cite{16} introduced a PCB defect detection method based on PSO-optimized threshold segmentation and SURF features. This approach utilized Particle Swarm Optimization (PSO) to improve the Otsu segmentation algorithm and combined SURF with FLANN for feature matching, further improving detection accuracy. Although these traditional algorithms are simple and fast, they are sensitive to illumination changes and environmental noise, resulting in poor robustness. Furthermore, they require manual tuning of numerous parameters, making it difficult to adapt to complex and diverse defect morphologies. With the advancement of computer technology, machine learning methods have gradually replaced traditional image processing. Machine learning approaches can automatically learn defect discrimination rules and feature patterns from training data, handling illumination changes and environmental noise more effectively without relying on manual parameter tuning.

Machine learning methods are mainly categorized into traditional machine learning (hand-crafted features + classifiers) and deep learning. Traditional machine learning includes Support Vector Machines (SVM)\cite{17}, Random Forests\cite{18}, and AdaBoost\cite{19}. Zhang et al.\cite{20} proposed a defect detection algorithm for surface-mounted devices based on Random Forest. By extracting shape, grayscale, and texture features from sub-regions for classification, the detection speed was significantly increased. Yao et al.\cite{21} developed a PCB defect detection technique based on Self-Supervised Learning with Local Image Patches (SLLIP). This method utilized relative position estimation of image patches, spatial adjacency similarity, and k-means clustering to learn semantic features, achieving excellent performance by detecting anomalies through feature differences. However, traditional machine learning methods rely heavily on hand-crafted features and cannot autonomously learn feature representations of PCB defects. Faced with complex and variable defect characteristics, hand-crafted features are not only time-consuming to design but also limited in expression capability and generalization. Consequently, deep learning methods, which can autonomously learn feature representations, have become the more widespread choice.

Deep learning methods are primarily divided into two architectures: one-stage and two-stage. Two-stage methods adopt a "region proposal-refinement classification" strategy, first generating candidate regions that may contain defects, and then performing classification and boundary regression on these regions. Representative algorithms include the R-CNN series (R-CNN, Fast R-CNN\cite{22}, Faster R-CNN\cite{23}) and Mask R-CNN\cite{24}. Liu et al. \cite{25}proposed a defect detection method based on the Swin Transformer, integrating it as the backbone into Cascade Mask R-CNN to achieve efficient PCB surface detection. Fontana et al.\cite{26} developed the open-source PCB dataset SolDef\_AI and implemented an innovative defect detection method using the Mask R-CNN algorithm. However, two-stage detectors suffer from high computational complexity and slow inference speeds, making it difficult to meet the real-time detection requirements of the industry. In contrast, one-stage deep learning methods employ an end-to-end unified structure, directly predicting target categories and locations from the input image. This approach significantly improves speed while maintaining accuracy, better aligning with the needs of industrial real-time detection.

One-stage deep learning methods mainly include SSD (Single Shot MultiBox Detector)\cite{27} and the YOLO (You Only Look Once) \cite{28}series. SSD represents an early one-stage detector that performs detection via multi-scale feature maps, extracting features at different levels to handle objects of various sizes. Wan et al. \cite{29} proposed a semi-supervised defect detection method with data expansion strategies (DE-SSD). By utilizing a batch addition strategy (BA-SSL) to reduce interference from unlabeled data and expanding the target dataset with labeled samples from other datasets, this method improved mAP by at least 4.7\% on the DeepPCB dataset compared to previous methods. Shi et al. \cite{30} introduced the Single Shot Detector with Rich Semantics (SSDT) for tiny PCB defect detection. This model fuses features from different levels through a semantic enhancement module and propagates deep semantics to shallow layers. Combined with attention mechanisms and a shuffle module to eliminate aliasing effects, it achieved 81.3\% mAP. However, SSD relies heavily on anchor mechanisms and employs simple feature fusion, often leading to unstable localization and recall for extremely tiny objects. The YOLO series has continuously evolved in feature fusion and detection head design. It is more friendly to small object detection while ensuring real-time performance, and possesses a mature ecosystem for engineering deployment. Therefore, YOLO is more widely applied in both academic research and industrial inspection.

\subsection{YOLO-based PCB Defect Detection}
As the representative of one-stage object detectors, the YOLO (You Only Look Once) series has been widely adopted in industrial inspection due to its excellent balance between speed and accuracy. From YOLOv5 to YOLOv11, continuous optimizations in backbone networks, feature enhancement and fusion, and detection heads have steadily improved detection performance. YOLOv5 introduced the Focus structure and CSP module; YOLOv7 \cite{31} proposed E-ELAN to achieve efficient feature aggregation; YOLOv8 adopted a decoupled head and an Anchor-Free design; and YOLOv11 further optimized the C3K2 module and detection efficiency. Although the latest YOLOv13 \cite{32} proposed a Hypergraph-based Adaptive Correlation Enhancement (HyperACE) mechanism, which uses a learnable hyperedge construction module to adaptively explore high-order correlations between vertices, this mechanism relies on sufficient vertex (pixel) information to build meaningful high-order relationships. In the context of PCB surface defect detection, the feature basis provided by extremely tiny defects is severely insufficient. Consequently, the input-output ratio of the HyperACE mechanism is extremely low, making it difficult to be effective. Therefore, this study adopts YOLOv11 as the baseline model and implements targeted improvements tailored to the specific characteristics of PCB defect detection.

In the field of PCB defect detection, researchers have carried out extensive improvements based on the YOLO framework. regarding attention mechanisms, Ming et al.\cite{33} proposed YOLO-HMC, which combines YOLOv5 with an improved Multi-Convolution Block Attention Module (MCBAM) to enhance the model's feature extraction capability by strengthening the interaction between channel and spatial features. Li et al.\cite{34} introduced C3F-SimAM-YOLO, integrating a C3-Faster-based SimAM parameter-free attention mechanism to infer 3D attention weights without increasing the parameter count. Yan et al.\cite{35} proposed a novel GS-YOLO network based on the YOLOv5s framework, designing a GA-SPPF module that extracts global information and combines it with local information from the SPPF module, helping the model learn both local and global features simultaneously. However, these attention mechanisms share a common limitation: most employ fixed receptive fields or generic multi-scale designs, failing to optimize for the specific scale distribution characteristics of tiny PCB defects.

Furthermore, some researchers have improved the YOLO model in terms of feature extraction, enhancement, and fusion to make it more suitable for PCB surface defect detection. Zhou et al.\cite{36} proposed TDD-YOLO, where the backbone network adopts a four-layer ME structure to integrate more low-level information and extract effective features. Mo et al. \cite{37} presented SE-ENv2 GC-Neck TSCODE (SGT) based on YOLOv5, integrating the SE-ENv2 backbone to retain more detail and positional information about tiny defects while emphasizing key features. He et al. \cite{38}proposed ABF-YOLO, introducing an embedded BiFusion module to facilitate multi-scale information fusion. However, these methods mainly focus on network architecture adjustments while neglecting the critical impact of loss functions and upsampling modules on tiny object detection. IoU-based loss functions are highly sensitive to positional deviations of tiny objects, leading to unstable gradients during training. Meanwhile, traditional interpolation-based upsampling methods tend to blur weak edge textures, causing tiny defect features to be lost before fusion.
In summary, although the aforementioned methods have improved PCB defect detection performance, deficiencies remain in detail preservation for tiny defects and targeted scale modeling, providing space for further research.

\section{Methodology}
\subsection{Overall Architecture of SME-YOLO}
\begin{figure}[!htbp]
    \centering
    \includegraphics[width=0.8\textwidth]{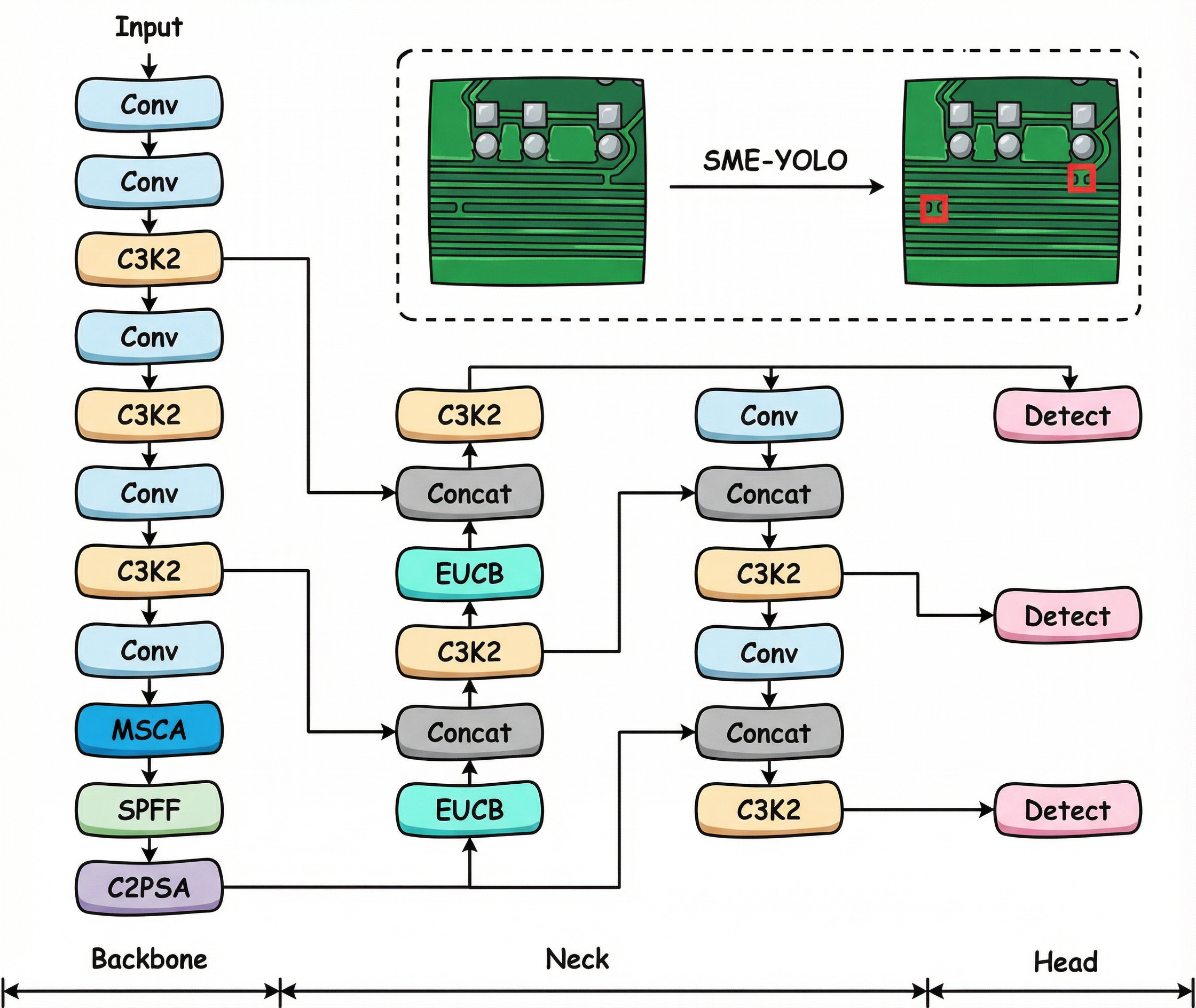} 
    \caption{SME-YOLO Overall Architecture.}
    \label{fig:fig1}
\end{figure}

YOLOv11 is a one-stage object detection model that continues the "end-to-end detection" design philosophy of the YOLO series. The model offers five versions of varying scales: n (nano), s (small), m (medium), l (large), and x (extra-large), to meet the requirements of different application scenarios. Considering the strict constraints on real-time performance and resource consumption in industrial PCB defect detection, this study selects YOLOv11n as the baseline model. Based on this, the proposed SME-YOLO retains the classic three-stage architecture of the YOLO series, comprising a Backbone, a Neck, and a Head. The complete network structure is illustrated in Figure \ref{fig:fig1}.

The Backbone is responsible for extracting multi-scale features. In this study, the C3k2 module in the 8th layer is replaced by the Multi-Scale Focused Attention (MSFA) module. The MSFA module perceives multi-scale features through parallel branches with different receptive fields. Its convolution kernel configuration is specifically optimized for the scale distribution of PCB defects, employing dual same-scale branches to form an implicit ensemble effect. The Neck handles feature fusion and information propagation. Here, the upsampling module is optimized into the Efficient Upsampling Convolution Block (EUCB). This module enhances feature expression and detail preservation while recovering spatial resolution, thereby improving the feature separability and recall rate for tiny defects. The Head serves as the final output layer of the network, mapping feature maps to detection results and outputting the location and category information of PCB defects. During the training process, NWDLoss replaces the original CIoU Loss. By modeling bounding boxes as Gaussian distributions and utilizing the Normalized Wasserstein Distance as a metric, NWDLoss improves the robustness of localization for tiny objects.

\subsection{Improved Content Description}
\subsubsection{NWDLoss: Normalized Wasserstein Distance Loss}

The IoU-based loss is widely used for bounding-box regression in object detection. Given a predicted box \(B_p=(x_p,y_p,w_p,h_p)\) and a ground-truth box \(B_g=(x_g,y_g,w_g,h_g)\), the IoU is defined as
\begin{equation}
\text{IoU}=\frac{|B_p\cap B_g|}{|B_p\cup B_g|}.
\end{equation}

\begin{figure}[!htbp]
    \centering
    \includegraphics[width=0.8\textwidth]{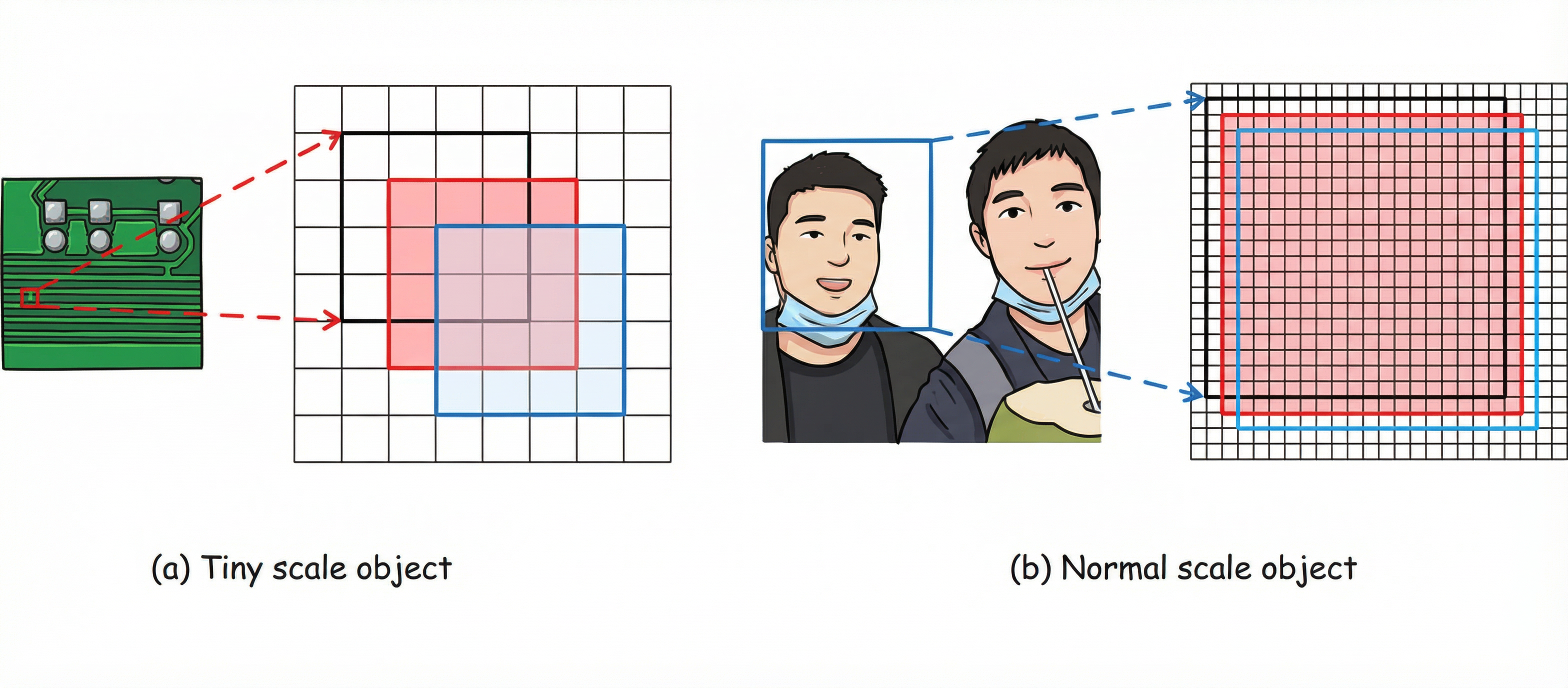} 
    \caption{Comparison of IoU sensitivity to misalignment for tiny targets and for conventional-sized targets.}
    \label{fig:fig2}
\end{figure}

However, IoU is highly sensitive to small target localization errors. As shown in Figure \ref{fig:fig2}, when the target is a normal-sized face, a tiny pixel-level offset in the predicted box slightly decreases IoU (from 0.81 to 0.66). For a very small target such as a PCB surface defect, the same offset can cause a dramatic drop (from 0.39 to 0.14), leading to unstable training and degraded localization accuracy for tiny defects.

To mitigate IoU’s sensitivity to small targets, we introduce NWDLoss. The core idea is to model a bounding box as a two-dimensional Gaussian distribution and measure the distance between distributions instead of using the traditional geometric IoU. For a bounding box \(B=(c_x,c_y,w,h)\), it is modeled as a 2D Gaussian distribution \(\mathcal{N}(\mu,\Sigma)\), where the mean vector \(\mu\) and the covariance matrix \(\Sigma\) are set to:
\begin{equation}
\mu=\begin{bmatrix} c_x \\ c_y \end{bmatrix},\qquad
\Sigma=\operatorname{diag}\left(\frac{w^2}{4},\,\frac{h^2}{4}\right).
\end{equation}

Based on this Gaussian modeling, the squared 2-Wasserstein distance between two bounding boxes is:
\begin{equation}
W_2^2\left(\mathcal{N}_p,\mathcal{N}_g\right)
= \|\mu_p-\mu_g\|_2^2 \;+\; \|\Sigma_p^{1/2}-\Sigma_g^{1/2}\|_F^2,
\end{equation}

where \(\|\cdot\|_F\) denotes the Frobenius norm.

To normalize the metric to the [0,1] interval, we define the Normalized Wasserstein Distance (NWD) as:
\begin{equation}
\text{NWD} = \frac{W_2^2\left(\mathcal{N}_p,\mathcal{N}_g\right)}{C},
\end{equation}

where \(C\) is a normalization constant related to the target scale.

The bounding-box regression loss is then defined as:
\begin{equation}
\mathcal{L}_{\text{NWD}} = 1 - \text{NWD}.
\end{equation}

Compared with IoU Loss, NWDLoss smooths the gradient fluctuations caused by small localization errors via distribution-distance measurement, leading to more stable training and improved localization accuracy for tiny defects.

\subsubsection{MSFA: Multi-Scale Focused Attention}
PCB surface textures exhibit high similarity, and defects are typically tiny, diverse, and vary significantly in scale. Traditional C3K2 modules primarily enlarge the receptive field through layer stacking. This approach can lead to the loss of fine-grained detail information for tiny objects during hierarchical transmission, weakening their edges and textures. Moreover, a single receptive field struggles to cater to the feature requirements of defects across different scales, thus limiting the model's adaptability to multi-scale defects. To address these issues, multi-scale attention mechanisms were proposed. These mechanisms form different receptive fields through branches with varying kernel sizes, allowing the model to dynamically select more appropriate scales for different defect sizes. However, their design often assumes a uniform distribution of target scales across various intervals. This assumption usually does not align with the actual scale distribution of datasets. Statistical analysis of the PKU-PCB dataset reveals that defect scales are not uniformly distributed; approximately 72.3\% of defects occupy 0.08\%-0.1\% of the original image area, exhibiting a clear dominant scale range. To tackle this, this study proposes a Multi-Scale Focused Attention (MSFA) module that focuses on dominant scales. Specifically, this module optimizes from the following three aspects:

\paragraph{Scale Focusing:} Adopts a focused convolution kernel configuration instead of a uniform one, specifically targeting the dominant scale range of the dataset, thereby strengthening feature extraction capabilities for critical scales.
\paragraph{Implicit Ensemble:} Retains a multi-branch parallel structure, utilizing branches with identical topology but independent parameters to achieve implicit model ensemble, enhancing the robustness and generalization ability of feature representation.
\paragraph{Computational Efficiency Optimization:} Removes redundant branches designed for large-scale targets. Given that PCB defects are predominantly tiny objects, large receptive field branches contribute little to detection accuracy. Their removal effectively reduces computational overhead while improving detection precision.

\begin{figure}[!htbp]
    \centering
    \includegraphics[width=0.8\textwidth]{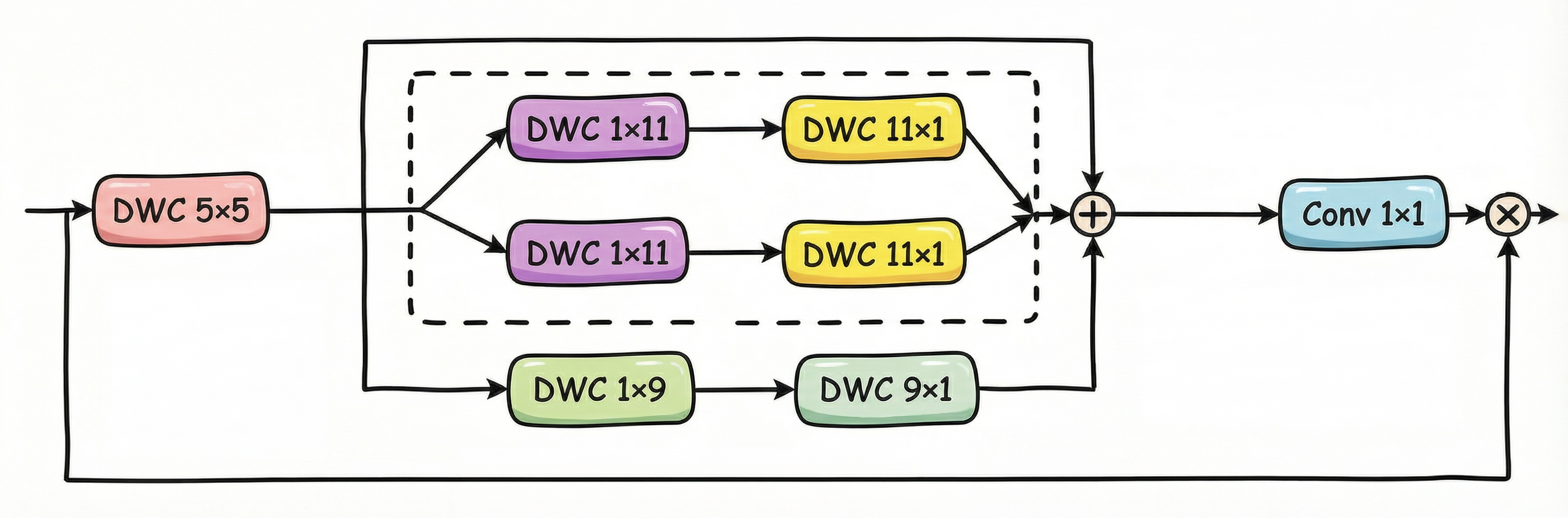} 
    \caption{Overall Structure of the MSFA Module.}
    \label{fig:fig3}
\end{figure}

The detailed structure of MSFA is shown in Figure \ref{fig:fig3}. The MSFA module consists of three core components: Focus-enhanced Multi-Scale Depthwise Convolution, Channel Mixing Layer, and Convolutional Attention Layer.

\paragraph{1) Focus-enhanced Multi-Scale Depthwise Convolution}
The MSFA module employs a three-branch parallel depthwise separable convolution structure. The convolution kernel configuration consists of two sets of dual branches (1×11 and 11×1) and one auxiliary branch (1×9 and 9×1). The two dual branches, having the same kernel size but independently learnable parameters, perform dual feature extraction for the dominant scale range of PCB defects. Although their topological structure is identical, independent parameter initialization and updates allow them to learn complementary feature representations, forming an implicit ensemble effect. The auxiliary branch retains a smaller scale to capture finer local features, complementing the 11×11 branches in terms of scale. Redundant large-scale branches are removed: considering that PCB defects are mainly tiny objects, large receptive field branches offer limited contribution to detection accuracy. Their removal significantly reduces computational cost.
\paragraph{2) Channel Mixing Layer}
After multi-branch feature extraction, the features are first concatenated along the channel dimension. Subsequently, a 1×1 convolution facilitates inter-branch channel information interaction and fusion:
\begin{equation}
F_{mixed}=Conv_{1\times1}(Concat(F_1,F_2,F_3,F_4))
\end{equation}

The channel mixing layer not only integrates feature information from different branches but also adaptively adjusts the contribution of each branch through learnable channel weights, enabling the model to dynamically balance the fusion ratio of multi-scale features based on the input content.
\paragraph{3) Convolutional Attention Layer}
To further enhance the model's ability to focus on critical defect regions, MSFA incorporates a lightweight convolutional attention mechanism. Spatial attention weight maps are generated via a 1×1 convolution, which then adaptively weights the fused features:
\begin{equation}
F_{output}=F_{input}\odot\sigma(F_{mixed})
\end{equation}

where \(\sigma\) is the Sigmoid activation function, and \(\odot\) denotes element-wise multiplication. This attention mechanism enables the model to adaptively strengthen the feature response in defect regions while suppressing background noise interference.

The proposed Multi-Scale Focused Attention module, focusing on dominant scales, achieves more efficient detection compared to traditional multi-scale attention mechanisms by concentrating computational resources on high-density scale regions. According to the research by Xie et al. [Reference] in ResNeXt, increasing the cardinality of parallel branches is more effective in improving model performance than simply increasing network depth or width. Although the dual branches have identical topological structures, their independent parameter spaces allow them to learn differentiated feature representations during training, creating an effect similar to model ensemble, which effectively enhances the model's detection performance.

\subsubsection{EUCB: Efficient Upsampling Convolution Block}
\begin{figure}[!htbp]
    \centering
    \includegraphics[width=0.8\textwidth]{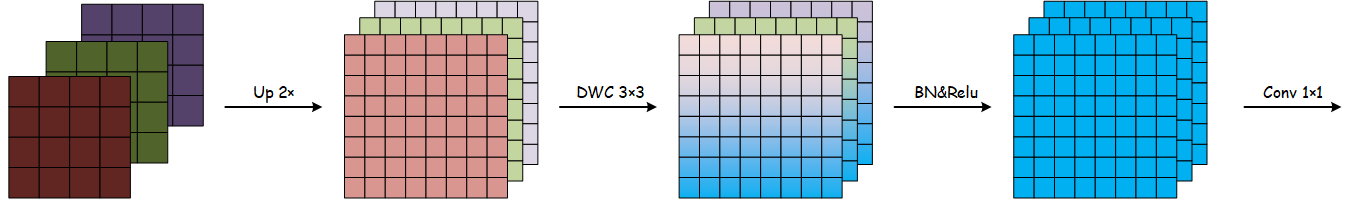} 
    \caption{Flowchart of the EUCB Module.}
    \label{fig:fig4}
\end{figure}

In the Neck network, upsampling of features is employed to restore the resolution of deep feature maps, facilitating fusion with high-resolution features from shallower layers. YOLOv11 utilizes bilinear interpolation for upsampling, a method that generates new pixel values by weighted averaging of neighboring pixels. This approach tends to smooth or dilute the edges of PCB surface defects, significantly degrading feature representation and severely impacting detection accuracy. To enhance the ability to restore defect edge details during upsampling, this paper introduces the Efficient Upsampling Convolution Block (EUCB). Its processing flow is illustrated in Figure \ref{fig:fig4}.

First, the resolution of the input feature map is expanded to double its original size to align with skip connections (skip features). Then, a 3×3 depthwise separable convolution is applied to the upsampled features for local neighborhood modeling and feature reorganization. This step mitigates the smoothing and diluting issues of PCB defect edges. Subsequently, the features undergo normalization and ReLU activation, which accelerate convergence while introducing non-linearity to enhance feature expression. Finally, a 1×1 convolution is performed to reduce the number of channels, matching the channel count of the next stage. The computational process is as follows:
\begin{equation}
EUC B(x) = C_{1\times 1}\left(\mathrm{ReLU}\left(\mathbf{LU}\left(\mathbf{BN}\left(\mathrm{DWC}\left(\mathrm{Up}(x)\right)\right)\right)\right)\right)
\end{equation}

\section{Experiments}
\subsection{Experimental Environment Setup}
All experiments were conducted on a Windows 10 platform, with the model implemented based on PyTorch 2.4.1 and Python 3.11.14. Both training and inference were performed in an environment equipped with an NVIDIA GeForce RTX 3060 Laptop GPU (6 GB VRAM) and CUDA 12.1. The CPU was an AMD Ryzen 7 5800H processor, and the system memory was 16GB. During the training phase, official pre-trained weights were loaded to accelerate convergence. The epoch count was set to 100, the batch size to 16, and an early stopping patience value of 60 was configured.

\subsection{Dataset}
\begin{figure}[!htbp]
    \centering
    \includegraphics[width=0.8\textwidth]{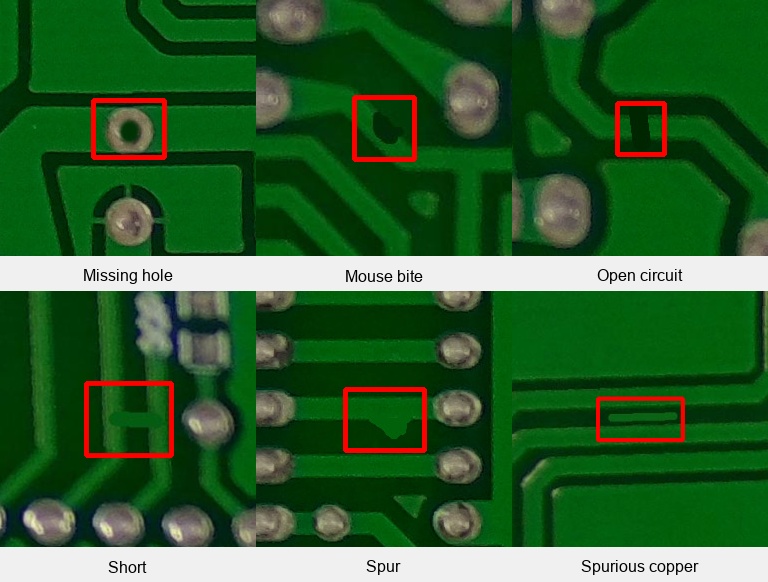} 
    \caption{Six Types of PCB Surface Defects.}
    \label{fig:fig5}
\end{figure}

This study validated its methodology using the PKU-Market-PCB dataset. This dataset, released by the Intelligent Robot Open Laboratory of Peking University, is specifically designed for PCB surface defect detection research. It comprises 1386 high-resolution PCB images, covering six common defect types: Missing hole, Mouse bite, Open circuit, Short, Spur, and Spurious copper, as illustrated in Figure \ref{fig:fig5}. Missing holes refer to blocked or undrilled vias that should be present, which can prevent component assembly. Mouse bites are small nicks on the board edge, which can lead to PCB breakage and failure over prolonged use. Open circuits are breaks in conductive traces that should be connected, directly causing functional failure of the PCB. Shorts occur when conductors that should be isolated become connected, potentially damaging components. Spurs are protrusions on the surface of the metal board, prone to leakage and short circuits. Spurious copper refers to unwanted excess copper foil or dots, easily leading to short circuits.

The sample counts, proportions, average areas, and average area proportions of the six defect types are presented in Table \ref{tab:table1}. Each defect category is uniformly distributed, accounting for approximately 16\% of the total, and each defect type occupies a minute area, roughly 0.1\% of the image area. The dataset is partitioned into training, validation, and test sets in an 8:1:1 ratio, ensuring mutual independence.
\begin{table}  
 \caption{Sample count, proportion, average area, and average area proportion of six defect types.}  
  \centering  
  \begin{tabular}{ccccc}  
    \toprule  
    Defect Type & Sample Count & Proportion & Average Area (pixels²) & Average Proportion of Image Area \\
    \midrule  
    Missing hole & 1194 & 17.03\% & 4620.16 & 0.08\% \\
    Mouse bite & 1176 & 16.77\% & 3895.58 & 0.07\% \\
    Open circuit & 1131 & 16.13\% & 2674.74 & 0.05\% \\
    Short & 1146 & 16.35\% & 7327.28 & 0.13\% \\
    Spur & 1173 & 16.73\% & 5047.77 & 0.09\% \\
    Spurious copper & 1191 & 16.99\% & 5427.00 & 0.09\% \\
    \bottomrule  
  \end{tabular}  
  \label{tab:table1}  
\end{table}  

\subsection{Evaluation Metrics}
To comprehensively assess the effectiveness of the proposed method, this study employs widely used standardized metrics: Precision (P), Recall (R), and mean Average Precision (mAP). These metrics are defined as follows:
\begin{equation}
P = \frac{TP}{TP + FP}
\end{equation}

\begin{equation}
R = \frac{TP}{TP + FN}
\end{equation}

\begin{equation}
mAP = \frac{\displaystyle \sum_{q=1}^{\Omega} AP(q)}{Q}
\end{equation}

where Precision represents the proportion of correctly predicted positive defect samples among all samples predicted as positive. Recall indicates the proportion of correctly detected positive samples among all true positive samples. mAP is obtained by averaging the Average Precision (AP) across all target categories, providing a comprehensive evaluation of the model's detection performance.

\subsection{Training Results}
\begin{figure}[!htbp]
    \centering
    \includegraphics[width=0.8\textwidth]{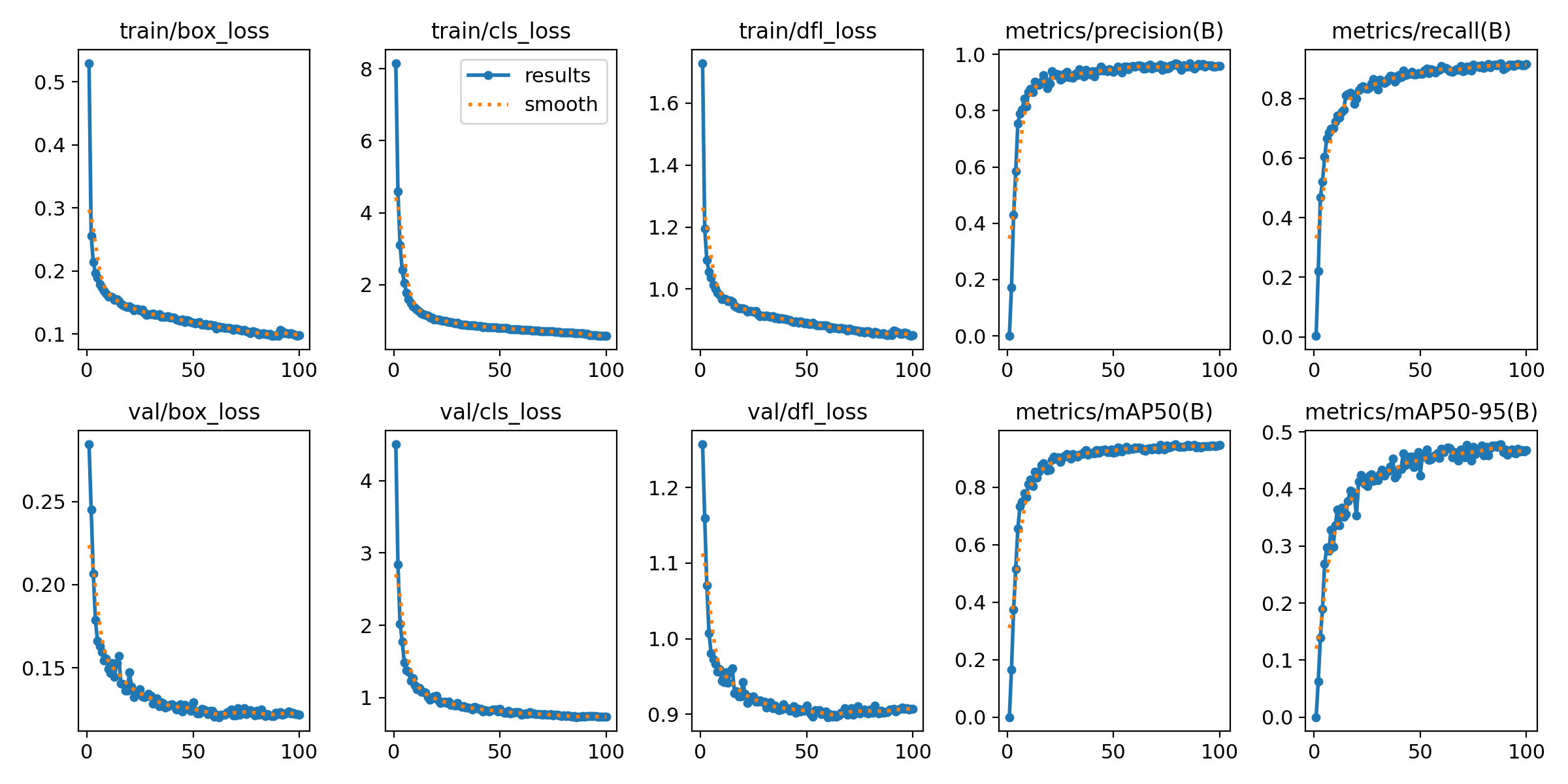} 
    \caption{Training Results of SME-YOLO.}
    \label{fig:fig6}
\end{figure}

After training the proposed SME-YOLO on the strictly partitioned PKU-Market-PCB dataset, the training and validation results are shown in Figure \ref{fig:fig6}. The loss value initially drops rapidly within the first 20-50 epochs, and the model quickly converges within 100 epochs. The overall trends for both the training and validation sets are largely consistent, eventually converging.

After testing the trained model and the baseline model on the same test set, the results are presented in Table \ref{tab:table2}. SME-YOLO demonstrates improved detection performance across all six defect types. With the exception of 'missing hole,' which has distinct features and is already close to saturation in the baseline model (mAP nearly 1.0), the mAP for all other detected defects shows an improvement of approximately 2\%.
\begin{table}[!htbp]
 \caption{Detection results of SME-YOLO and baseline model on the test set.}
  \centering
  \begin{tabular}{ccccccc}
    \toprule
    Model & Missing Hole & Mouse Bite & Open Circuit & Short & Spur & Spurious Copper \\
    \midrule
    YOLOv11 & 0.983 & 0.878 & 0.969 & 0.971 & 0.876 & 0.893 \\
    \textbf{SME-YOLO} & \textbf{0.984} & \textbf{0.897} & \textbf{0.986} & \textbf{0.992} & \textbf{0.928} & \textbf{0.913} \\
    \bottomrule
  \end{tabular}
  \label{tab:table2}
\end{table}

\subsection{Performance Comparisons}
\begin{figure}[!htbp]
    \centering
    \includegraphics[width=0.8\textwidth]{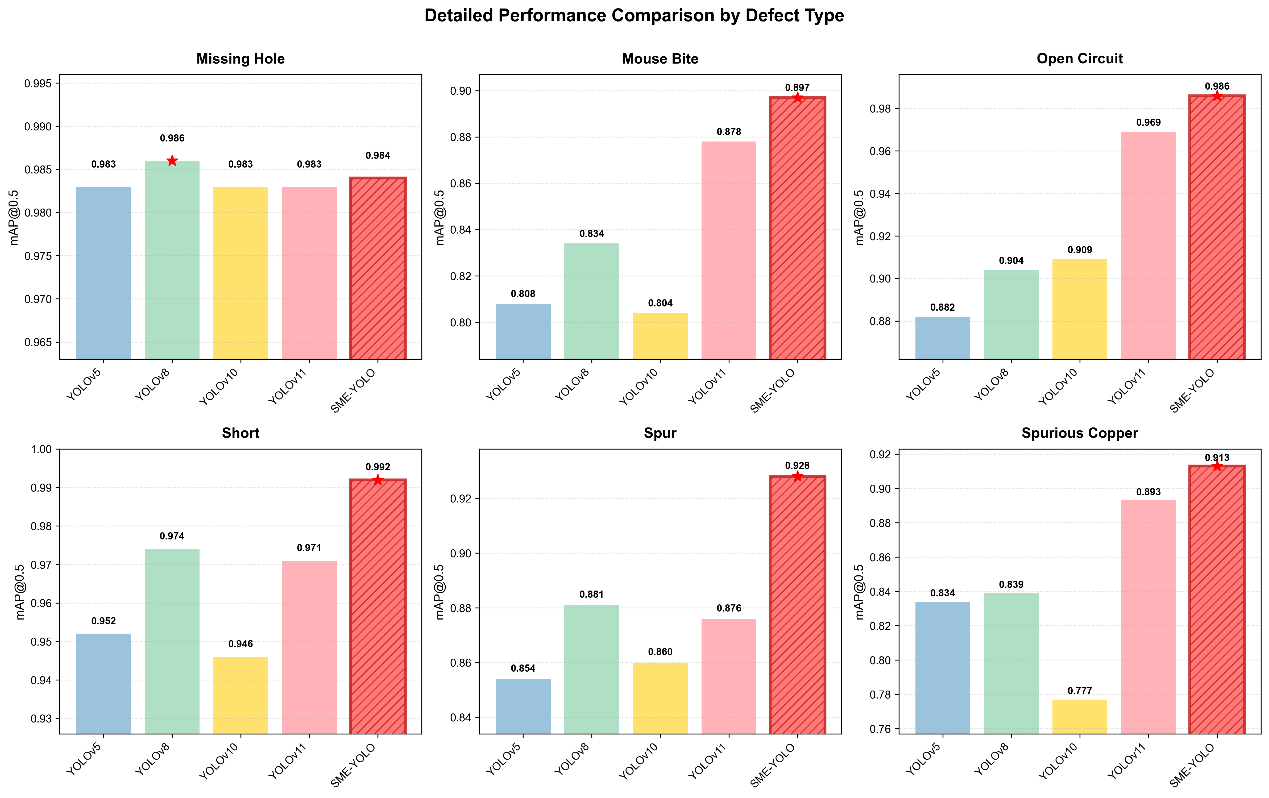} 
    \caption{Comparison of mAP@0.5 with state-of-the-art models on the six defect types.}
    \label{fig:fig7}
\end{figure}

To validate the superiority of the proposed SME-YOLO model, this study compares its detection performance with several other representative models under the same experimental environment. The results are shown in Table \ref{tab:table3} and Figure \ref{fig:fig7}. SME-YOLO achieves the highest mAP for five out of the six defect types, excluding 'missing hole,' which already reached near-saturation mAP.
\begin{table}[!htbp]
 \caption{Performance comparison of SME-YOLO with other representative models.}
  \centering
  \begin{tabular}{ccccccc}
    \toprule
    Model & Missing Hole & Mouse Bite & Open Circuit & Short & Spur & Spurious Copper \\
    \midrule
    YOLOv5 & 0.983 & 0.808 & 0.882 & 0.952 & 0.854 & 0.834 \\
    YOLOv8 & \textbf{0.986} & 0.834 & 0.904 & 0.974 & 0.881 & 0.839 \\
    YOLOv10 & 0.983 & 0.804 & 0.909 & 0.946 & 0.860 & 0.777 \\
    YOLOv11 & 0.983 & 0.878 & 0.969 & 0.971 & 0.876 & 0.893 \\
     \textbf{SME-YOLO} & 0.984 & \textbf{0.897} & \textbf{0.986} & \textbf{0.992} & \textbf{0.928} & \textbf{0.913} \\
    \bottomrule
  \end{tabular}
  \label{tab:table3}
\end{table}

\begin{figure}[!htbp]
    \centering
    \includegraphics[width=0.8\textwidth]{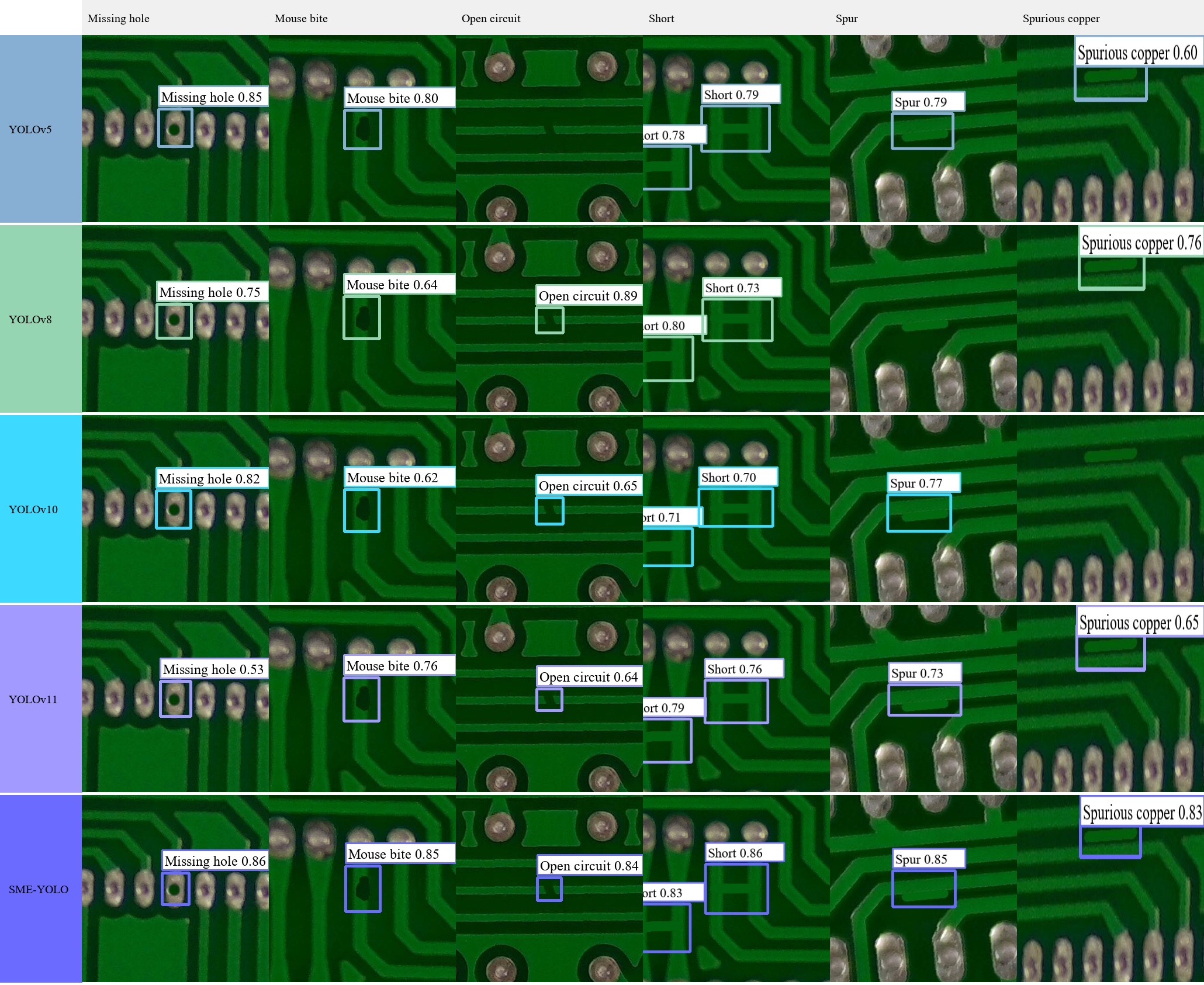} 
    \caption{Result Test Diagram for Ablation Experiments.}
    \label{fig:fig8}
\end{figure}

As shown in Table \ref{tab:4}, the SME-YOLO model exhibits overall superior performance. SME-YOLO achieves the highest mAP@0.5 (0.950), mAP@0.95 (0.480), Precision (0.970), and Recall (0.910) among all models. Compared to YOLOv11, these represent increases of 2.2\%, 0.2\%, 4\%, and 1.8\%, respectively.
\begin{table}[!htbp]
 \caption{Overall performance comparison of models.}
  \centering
  \begin{tabular}{ccccc}
    \toprule
    Model & P & R & mAP@0.5 & mAP@0.5:0.95 \\
    \midrule
    YOLOv5 & 0.943 & 0.817 & 0.885 & 0.425 \\
    YOLOv8 & 0.951 & 0.851 & 0.903 & 0.438 \\
    YOLOv10 & 0.888 & 0.801 & 0.880 & 0.425 \\
    YOLOv11 & 0.930 & 0.892 & 0.928 & 0.478 \\
    \textbf{SME-YOLO} & \textbf{0.970} & \textbf{0.910} & \textbf{0.950} & \textbf{0.480} \\
    \bottomrule
  \end{tabular}
  \label{tab:table4}
\end{table}
To further validate the effectiveness and superiority of the model, this study applied the aforementioned models to randomly selected PCB surface defect images. The results are shown in Figure \ref{fig:fig8}. Compared to other models, SME-YOLO provides more accurate defect region localization and segmentation, with higher confidence.

\subsection{Ablation Study}
To verify the contributions of the three improved modules—NWDLoss, EUCB, and MSFA—to the model's performance, this study conducted an ablation experiment on SME-YOLO. The results are presented in Table \ref{tab:table5}. When only the loss function on the baseline YOLOv11 was optimized to NWDLoss, mAP@0.5 improved from 0.928 to 0.939. This improvement is attributed to NWDLoss's greater robustness to small object boundary deviations compared to traditional IoU-based losses. It provides more stable gradients and more effective regression supervision when tiny objects are present, where even slight localization errors can cause IoU to fluctuate significantly, thereby enhancing the localization and detection of small defects. Upon introducing the EUCB upsampling module, mAP@0.5 further increased to 0.946 (an improvement of 0.7\%). This is because EUCB, through more effective feature upsampling and detail recovery, enhances the fidelity of edge and texture information in high-resolution features, improving multi-scale feature fusion quality. Consequently, it leads to a more comprehensive representation of tiny and weakly textured defects. With the additional integration of the MSFA attention module, the model achieved optimal performance in P, R, and mAP metrics (mAP@0.5 increased to 0.950, and mAP@0.5:0.95 to 0.480). MSFA adaptively emphasizes critical defect-related regions and scale information within multi-scale features, enhancing feature selectivity and cross-scale fusion efficiency. This results in a further gain in overall performance by reducing false positives while improving recall. Overall, the three improvements exhibit good complementarity, enabling the model to achieve superior detection results while maintaining low computational overhead.
\begin{table}[!htbp]
 \caption{Ablation study of SME-YOLO for the contribution of improved modules.}
  \centering
  \begin{tabular}{cccccc}
    \toprule
    Model & P & R & mAP@0.5 & mAP@0.5:0.95 & GFLOPs \\
    \midrule
    YOLOv11 & 0.930 & 0.892 & 0.928 & 0.478 & 6.3 \\
    +NWDLoss & 0.955 & 0.902 & 0.939 & 0.478 & 6.3 \\
    +NWDLoss+EUCB & 0.962 & 0.907 & 0.946 & 0.478 & 7.3 \\
    \textbf{+NWDLoss+EUCB+MSFA} & \textbf{0.970} & \textbf{0.910} & \textbf{0.950} & \textbf{0.480} & \textbf{7.1} \\
    \bottomrule
  \end{tabular}
  \label{tab:table5}
\end{table}

\section{Conclusion}
This study proposes SME-YOLO, a real-time object detector specifically designed for extremely tiny PCB surface defects. The introduced NWDLoss effectively mitigates the sensitivity of IoU to positional deviations in tiny objects. Subsequently, this study proposes the Multi-Scale Focused Attention (MSFA) module. By locking onto the main distribution scales of defects, MSFA effectively suppresses interference from irrelevant background noise. While ensuring lightweight design, it significantly enhances the model's ability to aggregate and represent multi-scale defect features. Finally, this paper introduces EUCB to replace the original upsampling module, using multi-scale convolutions to gradually recover edge and texture details of tiny objects. Compared to current popular PCB defect detection models, the proposed SME-YOLO demonstrates superior performance.

However, this study still has certain limitations. The experiments were solely validated on the PKU-PCB dataset, which has a relatively limited number of defect categories and sample scale. The model's generalization ability to larger-scale industrial datasets with more defect types requires further validation. Furthermore, although SME-YOLO achieved good performance on datasets with sufficient lighting, it was not systematically tested under more complex backgrounds and harsher imaging conditions, such as strong reflections, high dynamic lighting, or oil and dust contamination. Future research will focus on collecting data from real production lines covering more board types, more defect categories, and more working conditions to systematically evaluate the model's generalization ability. Concurrently, efforts will be made to further improve robustness and deployment usability in complex backgrounds, for instance, by introducing more targeted hard sample mining and data augmentation strategies.

\nocite{*}
\bibliographystyle{unsrt}
\bibliography{SME-YOLO.bib}

\begin{thebibliography}{10}

\bibitem{1}
Y.~Gao, Z.~Li, Y.~Wang, and S.~Zhu.
\newblock A novel yolov5\_es based on lightweight small object detection head for pcb surface defect detection.
\newblock {\em Scientific Reports}, 14(1):23650, October 2024.

\bibitem{2}
Yesong Wang, Binbin Wu, Lihua Zhang, Zhenyao Wang, Junwei Liu, Junjun Dong, and Jing Shi.
\newblock Enhanced pcb defect detection via hsa-rtdetr on rt-detr.
\newblock {\em Scientific Reports}, 15(1):31783, August 2025.

\bibitem{3}
F.~Yi, A.~S.~A. Mohamed, M.~H.~M. Noor, F.~Che Ani, and Z.~E. Zolkefli.
\newblock Yolov8-dee: a high-precision model for printed circuit board defect detection.
\newblock {\em PeerJ Computer Science}, 10:e2548, 2024.

\bibitem{4}
A.~Niaz, M.~Umraiz, S.~Soomro, and K.~N. Choi.
\newblock Vision transformer and mamba-attention fusion for high-precision pcb defect detection.
\newblock {\em PLOS ONE}, 20(9):e0331175, 2025.

\bibitem{5}
J.~Kim, J.~Ko, H.~Choi, and H.~Kim.
\newblock Printed circuit board defect detection using deep learning via a skip-connected convolutional autoencoder.
\newblock {\em Sensors}, 21(15):4968, July 2021.

\bibitem{6}
X.~Shen, Y.~Xing, J.~Lu, and F.~Yu.
\newblock Detection of surface defect on flexible printed circuit via guided box improvement in ga-faster-rcnn network.
\newblock {\em PLOS ONE}, 18(12):e0295400, 2023.

\bibitem{7}
Shengping Lv, Bin Ouyang, Zhihua Deng, Tairan Liang, Shixin Jiang, Kaibin Zhang, Jianyu Chen, and Zhuohui Li.
\newblock A dataset for deep learning based detection of printed circuit board surface defect.
\newblock {\em Scientific Data}, 11(1):811, July 2024.

\bibitem{8}
G.~Xiao, S.~Hou, and H.~Zhou.
\newblock Pcb defect detection algorithm based on cdi-yolo.
\newblock {\em Scientific Reports}, 14(1):7351, March 2024.

\bibitem{9}
R.~Sha, Z.~Zhang, X.~Cui, and Q.~Mu.
\newblock A lightweight cross-scale feature fusion model based on yolov8 for defect detection in sewer pipeline.
\newblock {\em PLOS ONE}, 20(8):e0330677, 2025.

\bibitem{10}
Q.~Zhao, S.~Liu, S.~Zhang, and B.~Wang.
\newblock Campus risk detection using the s-yolov10-sic network and a self-calibrated illumination algorithm.
\newblock {\em Sci Rep}, 15(1):24209, Jul 2025.

\bibitem{11}
M.~Zhang, Y.~Hu, B.~Xu, L.~Luo, and S.~Wang.
\newblock Dsf-yolo for weld defect detection in x-ray images with dynamic staged fusion.
\newblock {\em Scientific Reports}, 15(1):23305, July 2025.

\bibitem{12}
J.~Wang, C.~Xu, W.~Yang, and L.~Yu.
\newblock A normalized gaussian wasserstein distance for tiny object detection.
\newblock {\em arXiv}, June 2022.

\bibitem{13}
M.~M. Rahman, M.~Munir, and R.~Marculescu.
\newblock Emcad: Efficient multi-scale convolutional attention decoding for medical image segmentation.
\newblock {\em arXiv}, May 2024.

\bibitem{14}
{Intelligent Robot Open Lab, Peking University Shenzhen Graduate School}.
\newblock Pku-market-pcb dataset.
\newblock Shenzhen, China.
\newblock Accessed: 2024-11-15.

\bibitem{15}
A.-A. I.~M. Hassanin, F.~E. Abd El-Samie, and G.~M. El~Banby.
\newblock A real-time approach for automatic defect detection from pcbs based on surf features and morphological operations.
\newblock {\em Multimedia Tools and Applications}, 78(24):34437--34457, December 2019.

\bibitem{16}
Yuanpei Chang, Ying Xue, Yu~Zhang, Jingguo Sun, Zhangyuan Ji, Hewei Li, Teng Wang, and Jiancun Zuo.
\newblock Pcb defect detection based on pso-optimized threshold segmentation and surf features.
\newblock {\em Signal, Image and Video Processing}, 18(5):4327--4336, July 2024.

\bibitem{17}
C.~Cortes and V.~Vapnik.
\newblock Support-vector networks.
\newblock {\em Machine Learning}, 20(3):273--297, September 1995.

\bibitem{18}
L.~Breiman.
\newblock Random forests.
\newblock {\em Machine Learning}, 45(1):5--32, October 2001.

\bibitem{19}
Y.~Freund and R.~E. Schapire.
\newblock A decision-theoretic generalization of on-line learning and an application to boosting.
\newblock {\em Journal of Computer and System Sciences}, 55(1):119--139, August 1997.

\bibitem{20}
Wei Zhang, Yi~Lu, Tao Chen, and Jingwei Li.
\newblock A random forest algorithm for pcb smd defect detection.
\newblock {\em IEEE Transactions on Components, Packaging and Manufacturing Technology}, 15(5):1135--1142, May 2025.

\bibitem{21}
Naifu Yao, Yongqiang Zhao, Seong~G. Kong, and Yang Guo.
\newblock Pcb defect detection with self-supervised learning of local image patches.
\newblock {\em Measurement}, 222:113611, 2023.

\bibitem{22}
R.~Girshick.
\newblock Fast r-cnn.
\newblock {\em arXiv}, September 2015.

\bibitem{23}
S.~Ren, K.~He, R.~Girshick, and J.~Sun.
\newblock Faster r-cnn: Towards real-time object detection with region proposal networks.
\newblock {\em arXiv}, January 2016.

\bibitem{24}
K.~He, G.~Gkioxari, P.~Doll{\'a}r, and R.~Girshick.
\newblock Mask r-cnn.
\newblock {\em arXiv}, January 2018.

\bibitem{25}
Yulong Liu, Hao Wu, Youzhi Xu, Xiaoming Liu, and Xiujuan Yu.
\newblock Automatic pcb sample generation and defect detection based on controlnet and swin transformer.
\newblock {\em Sensors}, 24(11):3473, 2024.

\bibitem{26}
Gianmauro Fontana, Maurizio Calabrese, Leonardo Agnusdei, Gabriele Papadia, and Antonio Del~Prete.
\newblock Soldef\_ai: An open source pcb dataset for mask r-cnn defect detection in soldering processes of electronic components.
\newblock {\em Journal of Manufacturing and Materials Processing}, 8(3):117, 2024.

\bibitem{27}
Wei Liu, Dragomir Anguelov, Dumitru Erhan, Christian Szegedy, Scott Reed, Cheng-Yang Fu, and Alexander~C. Berg.
\newblock Ssd: Single shot multibox detector.
\newblock In {\em Computer Vision -- ECCV 2016}, volume 9905, pages 21--37, 2016.

\bibitem{28}
J.~Redmon, S.~Divvala, R.~Girshick, and A.~Farhadi.
\newblock You only look once: Unified, real-time object detection.
\newblock {\em arXiv}, May 2016.

\bibitem{29}
Yusen Wan, Liang Gao, Xinyu Li, and Yiping Gao.
\newblock Semi-supervised defect detection method with data-expanding strategy for pcb quality inspection.
\newblock {\em Sensors}, 22(20):7971, 2022.

\bibitem{30}
Wei Shi, Zhisheng Lu, Wei Wu, and Hong Liu.
\newblock Single-shot detector with enriched semantics for pcb tiny defect detection.
\newblock {\em The Journal of Engineering}, 2020(13):366--372, 2020.

\bibitem{31}
C.-Y. Wang, A.~Bochkovskiy, and H.-Y.~M. Liao.
\newblock Yolov7: Trainable bag-of-freebies sets new state-of-the-art for real-time object detectors.
\newblock In {\em 2023 IEEE/CVF Conference on Computer Vision and Pattern Recognition (CVPR)}, pages 7464--7475, Vancouver, BC, Canada, June 2023. IEEE.

\bibitem{32}
Mengqi Lei, Siqi Li, Yihong Wu, Han Hu, You Zhou, Xinhu Zheng, Guiguang Ding, Shaoyi Du, Zongze Wu, and Yue Gao.
\newblock Yolov13: Real-time object detection with hypergraph-enhanced adaptive visual perception.
\newblock {\em arXiv}, September 2025.

\bibitem{33}
Minghao Yuan, Yongbing Zhou, Xiaoyu Ren, Hui Zhi, Jian Zhang, and Haojie Chen.
\newblock Yolo-hmc: An improved method for pcb surface defect detection.
\newblock {\em IEEE Transactions on Instrumentation and Measurement}, 73:1--11, 2024.

\bibitem{34}
D.~Li, F.~Bai, S.~Li, B.~Liu, and L.~He.
\newblock El-pcbnet: An efficient and lightweight network for pcb defect detection.
\newblock {\em Measurement}, 253:117719, September 2025.

\bibitem{35}
G.~Li, Y.~Gan, W.~Zhang, and H.~Che.
\newblock Gs-yolo: A lightweight and high-performance method for pcb surface defect detection.
\newblock {\em Expert Systems with Applications}, 303:130583, March 2026.

\bibitem{36}
Wen Zhou, Changyi Li, Zhiwei Ye, Qiyi He, Zhe Ming, Jingliang Chen, Fang Wan, and Zhenhua Xiao.
\newblock An efficient tiny defect detection method for pcb with improved yolo through a compression training strategy.
\newblock {\em IEEE Transactions on Instrumentation and Measurement}, 73:1--14, 2024.

\bibitem{37}
C.~Mo, Z.~Hu, J.~Wang, and X.~Xiao.
\newblock Sgt-yolo: A lightweight method for pcb defect detection.
\newblock {\em IEEE Transactions on Instrumentation and Measurement}, 74:1--11, 2025.

\bibitem{38}
Xiaoyao He and Mingyang Xie.
\newblock Abf-yolo: A fine-grained pcb defect detection framework integrating axial attention and bidirectional feature fusion.
\newblock {\em IEEE Transactions on Consumer Electronics}, 71(3):8562--8570, 2025.

\end{thebibliography}

\end{document}